\pdfoutput=1

\documentclass[11pt]{article}

\usepackage{newunicodechar}

\usepackage[final]{acl}

\usepackage{times}
\usepackage{latexsym}

\usepackage[T1]{fontenc}

\usepackage[utf8]{inputenc}

\usepackage{microtype}

\usepackage{inconsolata}

\usepackage{graphicx}
\usepackage{booktabs}
\usepackage{multirow}
\usepackage{array}
\usepackage{geometry}
\usepackage{lscape}
\usepackage{caption}
\usepackage{ragged2e}
\usepackage[most]{tcolorbox}
\usepackage[dvipsnames]{xcolor}
\usepackage{setspace}

\usepackage{textcomp, gensymb}
\usepackage{amsmath}
\usepackage{stfloats}
\usepackage{subcaption}
\usepackage{enumitem}
\usepackage{tabularx,booktabs,seqsplit,caption}
\newcolumntype{Y}{>{\raggedright\arraybackslash}X}
\usepackage[table]{xcolor}  
\definecolor{rowgray}{gray}{0.95} 

\usepackage{amsmath,amssymb}
\usepackage{hyperref}

\title{LAVA: Logic-Aware Validation and Augmentation Framework for Large-Scale Financial Document Auditing}

\author{ \textbf{Ruoqi Shu}\thanks{Equal contribution.}\thanks{Corresponding author.}, \textbf{Xuhui Wang}\footnotemark[1]\footnotemark[2], \textbf{Isaac Wang}\thanks{Work done during an internship at BMO Financial Group.}, \textbf{Yanming Mai}, \textbf{Bo Wan} \\ \\ BMO Financial Group \\ \texttt{\{\href{mailto:ruoqi.shu@bmo.com}{ruoqi.shu}, \href{mailto:xuhuieric.wang@bmo.com}{xuhuieric.wang}\}@bmo.com}, \\ \texttt{\{\href{mailto:isaac.wang@bmo.com}{isaac.wang}, \href{mailto:yanming.mai@bmo.com}{yanming.mai}, \href{mailto:bo.wan@bmo.com}{bo.wan}\}@bmo.com} }

\begin{document}
\maketitle

\begin{abstract}
Financial document validation in production---such as payroll auditing, tax compliance, and loan underwriting---demands exceptional accuracy, consistency, and reproducibility under strict enterprise constraints. In practice, documents arrive with heterogeneous layouts and formats, semantically rich, context-dependent content, and embedded business rules that current pipelines struggle to process reliably. We introduce LAVA (Logic-Aware Validation and Augmentation)---a modular, backbone-agnostic pipeline built on multimodal large language models---that integrates a four-stage design: document–rule retrieval, layout-preserving information extraction, auxiliary metadata enrichment, and auditable symbolic/arithmetic verification. LAVA supports robust rule grounding, fine-grained error attribution, and consistent, traceable end-to-end execution—capabilities essential for high-stakes deployment. Evaluated on a large real-world benchmark with diverse financial documents and dozens of expert-curated validation rules, LAVA outperforms baselines in hallucination control and edge-case handling while maintaining efficient token usage, demonstrating practicality for high-volume, time-critical validation.
\end{abstract}

\section{Introduction}
Regulatory penalties, financial losses, and reputational damage can all result from a single error in financial document validation, making accuracy, consistency, and auditability non-negotiable. Financial institutions process millions of documents daily across workflows such as loan underwriting, payroll auditing, tax compliance, and fraud detection. The challenge is acute for semi-structured documents like statements, invoices, and tax slips, which span multiple pages, exhibit irregular layouts, encode domain-specific business logic, and often arrive as noisy scans or non-standard PDFs~\cite{bhattacharyya-etal-2025-information, ding2024deep, Xu_2020, chen2024rodla}.

Recent years have witnessed rapid progress in visually rich document understanding through layout- and structure-aware pretraining of multimodal large language models (MLLMs). LayoutLM~\cite{Xu_2020} pioneered spatial–textual joint encoding, followed by DocFormer~\cite{appalaraju2021docformerendtoendtransformerdocument}, DocLLM~\cite{wang-etal-2024-docllm}, mPLUG-DocOwl2~\cite{hu-etal-2025-mplug}, and ROP~\cite{zhang2024modelinglayoutreadingorder}, advancing multimodal architectures for structural representation.
This shift moves beyond pure text modelling to multimodal document intelligence. However, evaluations remain dominated by perceptual and question answering (QA) tasks, probing reasoning in a narrow, task-specific manner while rarely accessing validation---reasoning that tests both interpretation and reliability under structured and cross-field constraints.

Enterprise-grade document validation presents requirements beyond those of perception or reasoning alone. It calls for symbolic rule enforcement, cross-field consistency, and multi-step logical coherence---capabilities only partially reflected in existing benchmarks~\cite{Wang_2023,li2025gdibenchbenchmarkgeneraldocument,borchmann2021due}. Recent datasets have expanded evaluation to layout-aware perception and understanding~\cite{zhu2024mmdocbenchbenchmarkinglargevisionlanguage, wu2023dcqadocumentlevelchartquestion, mathew2021docvqadatasetvqadocument, Stanis_awek_2021, šimsa2023docilebenchmarkdocumentinformation}, but compliance-critical validation logic remains underexplored. In practice, enterprises often patch the gap by pairing general-purpose models with rigid rule-based modules~\cite{shende2024enhancing}. Yet even state-of-the-art models that excel at extraction or understanding exhibit ongoing limitations when directly applied to validation: hallucinations remain common, and reasoning traceability is limited. Adaptation across regulatory schemas is fragile, and computational costs escalate with token usage at enterprise scale. These shortcomings make validation not merely an extension but a distinct and emerging frontier of document intelligence—one that demands frameworks where accuracy, efficiency, audit readiness, and robustness are central requirements.

\paragraph{Our Work.}
To address these challenges, we present \textbf{LAVA} (\textbf{L}ogic-\textbf{A}ware \textbf{V}alidation and \textbf{A}ugmentation), a modular and efficient framework for verifiable reasoning over semi-structured, layout-complex financial documents, designed for real-world applicability, fine-grained error attribution, and rapid adaptation and reproducibility across structurally similar collections. Agnostic to backbone models, LAVA extends beyond static benchmarks by integrating (i) layout-informed knowledge extraction preserving structural cues, (ii) domain-aware augmentation with contextual metadata, and (iii) arithmetic and symbolic verification ensuring factual alignment with business rules, orchestrated in a four-stage system of retrieval, extraction, augmentation, and hybrid reasoning.

We evaluate LAVA on a real-world large-scale industrial benchmark with validation rules curated by senior industry experts reflecting real regulatory conditions. Results show competitive gains in factual accuracy and symbolic correctness, together with lower computational overhead than baseline MLLM pipelines, demonstrating robustness and cost-effectiveness in realistic financial validation scenarios.

Our main contributions are:
\begin{itemize}
\item \textbf{Task and System.} We formalize \emph{financial document validation} as a multi-document reasoning task—largely absent in existing benchmarks—and instantiate it in \textbf{LAVA}, a novel modular framework designed for accurate and auditable validation of semi-structured financial documents.

\item \textbf{Reasoning Strategy \& Auditability.} We design a controllable hybrid reasoning framework unifying factual/contextual templates with symbolic and arithmetic tasks via explicit formula generation, with an external checker fallback ensuring correctness, thereby enhancing accuracy, interpretability, and operational robustness.

\item \textbf{Evaluation.} We propose a comprehensive evaluation framework covering symbolic correctness, factual alignment, and hallucination control for fine-grained reasoning assessment in realistic validation workflows.
\end{itemize}

\section{Related Work}
\paragraph{Visually Rich Document Understanding.}
Recent work has shifted from extraction pipelines toward LLM-centric modeling that integrates layout and visual cues. DocLayLLM~\cite{liao2025doclayllmefficientmultimodalextension} adds visual patches and 2D positional tokens, VisDoM~\cite{suri-etal-2025-visdom} combines multimodal retrieval with consistency constraints, 3MVRD~\cite{ding-etal-2024-3mvrd} aligns fine- and coarse-grained signals via multi-task distillation, and LayoutLLM~\cite{fujitake-2024-layoutllm} applies instruction tuning for unified document tasks. These advances improve representation and generalization; our work is complementary, focusing on \emph{auditable validation}—explicit symbolic checks, cross-field consistency, and reproducible reasoning traces required in compliance-critical workflows. Our framework is model-agnostic, plugging into any MLLM backbone to introduce validation as a controllable layer.

\paragraph{Layout-Guided Document Encoding.}
Backbones such as LayoutLMv3~\cite{huang2022layoutlmv3pretrainingdocumentai}, FormNet~\cite{lee2022formnetstructuralencodingsequential, lee2023formnetv2multimodalgraphcontrastive}, and DocFormer~\cite{appalaraju2021docformerendtoendtransformerdocument} combine textual, spatial, and visual cues through large-scale pre-training. These excel at form-style entity extraction but remain embedding-level and not optimized for symbolic validation. In contrast, our framework leverages layout-derived structures to enable modular business rule checks and reproducible logic tracing, addressing auditability without massive training effort.

\paragraph{LLM Verification and Rule-based Validation.}
Progress in LLM factuality---via self-checking~\cite{dhuliawala-etal-2024-chain}, retrieval-augmented prompting~\cite{qin2025dontlethallucinatepremise}, symbolic grounding~\cite{torroba2023symgen}, and rectification~\cite{kang2023ever}---has improved reliability in clean text and formal reasoning~\cite{liu2025safe}. Yet these methods falter on noisy, heterogeneous documents with long-range dependencies and embedded business rules. Traditional rule engines (e.g., Drools) provide transparency but fail under layout variation, while hybrid LLM–rule or knowledge graph systems~\cite{vertsel2024hybrid, sadowski2025structured} trade flexibility for interpretability. Our framework instead couples symbolic verification with layout- and metadata-informed prompting, merging rule transparency with neural adaptability, making it fit for compliance-critical validation.

\section{Method}
\begin{figure*}[t]
    \centering
    \includegraphics[width=1\textwidth]{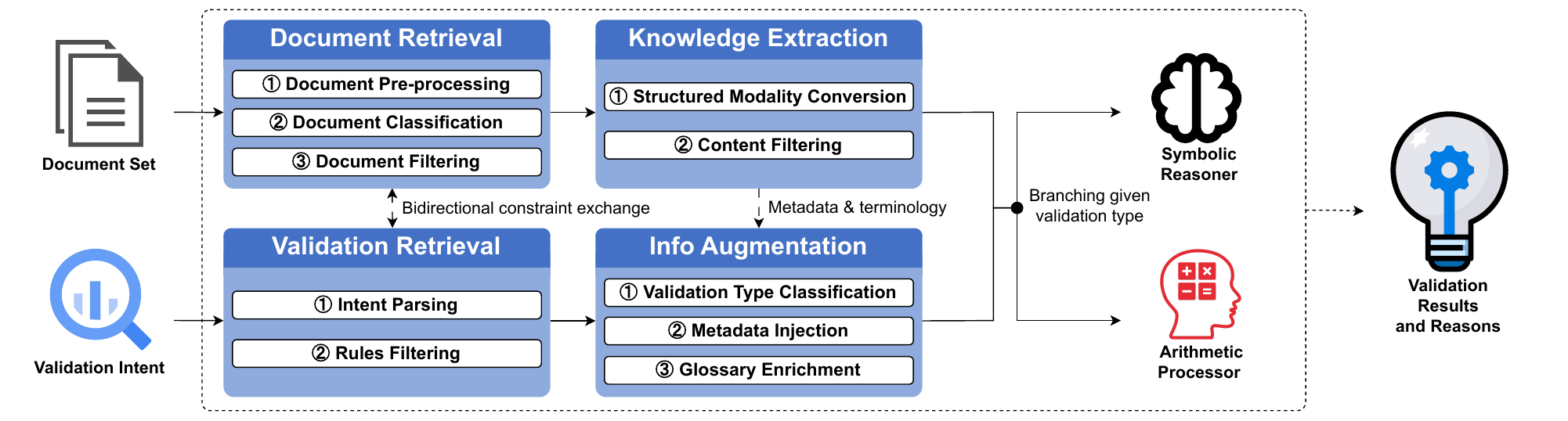}
    \vspace{-2em}
    \caption{Overview of LAVA. The architecture comprises two parallel pipelines---document processing and rule grounding---interacting via bidirectional constraints (solid arrows: main data flow; dashed arrows: auxiliary exchange). The document-processing track performs retrieval, extraction, and augmentation to produce layout-preserving structured content with enriched metadata. The rule-grounding track retrieves, classifies, and dispatches applicable validation rules to either the Symbolic Reasoner or Arithmetic Processor, depending on reasoning type.}
    \label{fig:overview}
\end{figure*}

\paragraph{Problem Formalization.}
We define \textbf{financial document validation} as a human-in-the-loop copilot task, where the system assists users (e.g., underwriters, compliance officers, fraud analysts) in verifying whether a set of financial documents (e.g., from a mortgage application) satisfies certain business rules under real-world conditions of layout complexity, domain-specific logic, and noisy input.

\paragraph{Inputs.}
The system takes as input:
(1) a document set $\mathcal{D} = \{d_1, \dots, d_N\}$ in scanned PDF or image format, where each $d_i$ is a semi-structured financial document (e.g., bank statement, tax form), and  
(2) a validation intent $q$, a user-specified verification goal in natural language (e.g., Does gross income exceed loan threshold?).

\paragraph{Objective.}
The system maps $(q, \mathcal{D})$ to a set of validation outputs $\mathcal{V} = \{v_k\}$, where each $v_k$ includes:
a subset of supporting documents $\mathcal{D}_k \subseteq \mathcal{D}$, 
a set of retrieved business rules $\mathcal{R}_k \subseteq \mathcal{R}$ from a \textbf{predefined} rule library, 
a binary verification label $y_k \in \{\texttt{Pass}, \texttt{Fail}\}$, and 
an explanation trace $e_k$ for auditability. This formulation supports multi-document, logic-grounded reasoning while ensuring modular, interpretable validation aligned with enterprise workflows.

\paragraph{Architecture.}
Figure ~\ref{fig:overview} illustrate LAVA architecture, which employs a modular design paradigm to address the complexity of enterprise-grade production systems.
Although this design involves the integration of multiple components, it yields critical advantages for deployment. Decoupling the workflow isolates potential points of failure within individual modules, enabling targeted and independent validation and thereby systematically reducing the long-term verification cost and overall operational overhead. Furthermore, the ability to debug, update, or replace modules without systemic disruption enhances maintainability---a stark contrast to the challenges of managing opaque, end-to-end models. This design philosophy is therefore foundational to building a robust, auditable, and scalable system fit for the rigors of real-world financial validation.

\subsection{Document and Validation Retrieval} \label{sec:docrec}

We jointly describe the first two modules, as they operate in a tightly coupled fashion to determine relevant document-rule pairs for downstream extraction.
Given a user-specified \textit{validation intent} $q$ and a document set $\mathcal{D} = \{d_1, \dots, d_N\}$, the modules select a subset $\mathcal{D}_v \subseteq \mathcal{D}$ that is temporally valid and relevant to the task, along with a set of executable business rules $\mathcal{R}_v = \{r_1, \dots, r_M\}$. Both subsets are tailored to the verification goal through a bidirectional constraint mechanism, ensuring only applicable document-rule pairs are forwarded for knowledge extraction and augmentation.

\paragraph{Document Retrieval.}
Documents are first preprocessed to normalize layout and correct visual artifacts (e.g., OCR errors, rotation, skew)~\cite{boudraa2020using}, preserving alignment and structural fidelity for downstream modules. Each document $d_i$ is then classified into a predefined document type using a lightweight image-based classifier such as TinyViT~\cite{wu2022tinyvitfastpretrainingdistillation}, as financial documents of the same type generally share consistent page-level features. Recognized types are forwarded to Validation Retrieval to constrain rule applicability.

Document-level metadata (e.g., date, coverage period) is extracted using a template-guided NER pipeline with regex patterns and rule-based heuristics. Temporal constraints parsed from the validation intent in the next module (e.g., ``past 3 months'') are applied to filter out documents outside the relevant time window. In addition, applicable document types extracted from retrieved rules in Validation Retrieval are passed back to further prune documents irrelevant to all candidate rules.

\paragraph{Validation Retrieval.}
Given $q$, this module retrieves a subset of rules $\mathcal{R}_v$ from the predefined library $\mathcal{R}$, based on semantic relevance and document compatibility. The rule library is enriched with metadata specifying applicable document types. Lightweight LLMs can parse $q$ to extract temporal constraints, while a sentence encoder (e.g., Sentence-BERT~\cite{reimers2019sentencebertsentenceembeddingsusing}) encodes $q$ to retrieve top-$K$ semantically relevant rules. Retrieved rules are then filtered using document-type constraints from Document Retrieval, and their own document-type metadata is fed back to further refine $\mathcal{D}_v$.

In conclusion, temporal and semantic cues from $q$ filter the document set, while recognized document types and rule metadata eliminate inapplicable rules. This closed-loop filtering minimizes irrelevant candidates on both sides, reduces reasoning load, and improves accuracy without sacrificing interpretability, ensuring downstream processing operates on the most relevant and valid document-rule pairs.


\subsection{Knowledge Extraction} \label{sec:ke}

This module transforms each filtered document $d_i \in \mathcal{D}_v$ into a compact, layout-aware hybrid representation for downstream reasoning. 
Instead of flat key–value pairs, we produce a structured markup that encodes page layout, visual grouping, and field dependencies, serving as a bridge between scanned formats and language-model-friendly input.

\paragraph{Structured Modality Conversion.}
To capture the rich visual and semi-structural semantics of financial documents, we adopt an HTML-like markup constructed from parsed layout and OCR signals.  
Prior work shows that retaining tabular alignments, hierarchical sections, and field groupings improves reasoning fidelity~\cite{sui2024table}, but raw markup is insufficient for noisy scans.  
We therefore augment it with:  
(1) structural parsing via document analysis tools (e.g., LayoutParser~\cite{shen2021layoutparser}, LayoutLMv3~\cite{huang2022layoutlmv3pretrainingdocumentai}), AWS Textract;  
(2) OCR-based recovery (e.g., Tesseract OCR~\cite{TessOverview}) for free-form or scattered content, including reconstruction of long paragraphs into coherent spans;  
(3) proximity-based grouping to merge fragmented tokens into coherent semantic units; and  
(4) visual region preservation for inherently non-textual content (e.g., charts, stamps, signatures), where candidate regions are identified from layout cues (e.g., low text coverage or OCR confidence) and retained as image patches for the MLLM input.

\paragraph{Content Filtering.}
The extracted representation often contains much noise from headers, footers, boilerplate blocks, or placeholders.  
We prune such elements using DOM structure, field labels (e.g., \texttt{Name}, \texttt{Address}), and positional cues, reducing token usage while improving attention focus for model prompts.

By combining structure-preserving markup, semantic recovery, selective visual preservation, and noise reduction, this module delivers a high-fidelity hybrid form that maintains interpretability while enabling reliable downstream reasoning.

\subsection{Information Augmentation} \label{sec:ia}

This module enriches downstream reasoning by injecting auxiliary signals from both documents and retrieved rules. It serves two purposes: routing rules to the appropriate verification pathway and augmenting prompts with rich metadata to improve model understanding.

Each rule $r_k \in \mathcal{R}_v$ is classified as symbolic (context-dependent logic) or arithmetic (numeric computation) via a lightweight LLM query, avoiding brittle heuristics. In parallel, metadata is additionally extracted from layout-preserving outputs: language (via token-based detection), document types (from retrieval module), and domain-specific terms (e.g., withholding or CPP~\footnote{CPP refers to the Canada Pension Plan, a mandatory pension contribution in Canadian payroll systems.}) detected through lexical and 
structural heuristics targeting low-frequency tokens, abbreviations, and left-hand-side predicates in field labels or rule expressions. This metadata is distilled into concise semantic clarifications for interpretability and disambiguation.  
The resulting signals are incorporated into prompt headers or reference blocks, sharpening alignment with rule semantics and improving precision in complex validation scenarios, while keeping the verification loop efficient.

\begin{figure*}[t]
    \centering
   \includegraphics[width=1\textwidth]{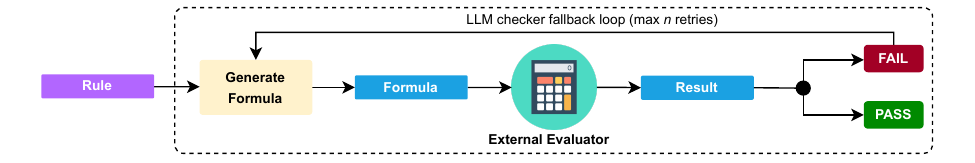}
    \vspace{-2em}
    \caption{Arithmetic Processor. Given a validation rule, the system generates a formula and delegates evaluation to an external calculator. A checker LLM validates alignment between the rule and result. Failures trigger a fallback loop with negative feedback.}
    \label{fig:computation}
\end{figure*}

\subsection{Validation}

To support diverse verification needs and enhance the reliability, we adopt a bifurcated framework with two sub-modules: \textbf{Arithmetic Processor} for numerical tasks (e.g., verifying tax deductions, calculating gross revenue), and \textbf{Symbolic Reasoner} for all other general rules requiring semantic and contextual reasoning.

\paragraph{Arithmetic Processor.}
As illustrated in Figure~\ref{fig:computation}, when rules involving arithmetic or numerical computation are routed to this sub-module, instead of relying on LLMs for direct computation---prone to hallucinations and numeric instability, we adopt a tool-use paradigm where the model is used solely for generating a task-specific formula, which is then executed by a deterministic external engine such as a Python interpreter~\cite{pal, pot}. To ensure alignment between the formula and the validation semantics, we introduce a fallback auditing loop: the rule and generated formula are reviewed by a secondary LLM "checker". If a mismatch is detected, the formula is regenerated, conditioned on the previous (incorrect) version as a negative example. This loop improves robustness by systematically detecting and correcting errors, mitigating hallucinated computations and flawed reasoning pathways.

\paragraph{Symbolic Reasoner.}
This sub-module handles semantic or structural reasoning. In contrast to Arithmetic Processor, it directly delegates rules to a general-purpose model, as such reasoning falls within the model's inherent strengths. This design choice reduces integration overhead and inference latency, while remaining sufficient for a wide range of non-arithmetic verification scenarios.

\paragraph{Prompts.}

We adopt our prompt construction strategy in meta-prompting fashion~\cite{metaprompting} that presents task-relevant information in a step-wise and zero-shot format rather than relying on illustrative examples. Unlike chain-of-thought~\cite{chainofthought}, few-shot prompting~\cite{fewshot}, or self-consistency~\cite{selfconsistency}, our approach avoids brittle curated examples, which struggle to generalize across heterogeneous financial documents~\cite{leasttomost}, and removes high inference cost of example-heavy prompting. 

\begin{figure}[t]
\centering
\input{prompt_box}
\vspace{-5mm}
\caption{
Prompt excerpt for Symbolic Reasoner, dynamically built from the retrieved validation rule, extracted knowledge, and augmentation components. 
Full versions for both Arithmetic Processor and Symbolic Reasoner are provided in Appendix~\ref{sec:prompt1} and \ref{sec:prompt2}.
}
\label{fig:simplified_prompt}
\end{figure}

To ensure verification prompts are precisely tailored to both the rules and associated documents, we implement a dynamic, template-based prompt construction system. It integrates supplemental information from upstream modules with the rule to populate flexible prompt templates. By leveraging templating libraries like Jinja, we employ programmatic, code-like logic to govern the inclusion and granularity of metadata based on both document layout and task semantics. As illustrated in Figure~\ref{fig:simplified_prompt}, this mechanism produces prompts that are maximally informative while remaining token-efficient and robust to both arithmetic and symbolic reasoning tasks. By structuring all inputs according to rule logic, metadata, and knowledge, our approach generalizes well across heterogeneous financial documents. The resulting system remains accurate, cost-efficient and context-aware, supporting scalable and reliable verification tasks.

\section{Experiments}

We evaluate LAVA on real-world Canadian mortgage application documents sampled from a proprietary database to assess the system within a consistent validation scenario. The set includes multiple document types and around 1,000 scanned PDFs or images---such as tax forms, bank/investment statements, and legal agreements---selected for their semi-structured formats, diverse layouts, and rich logical dependencies (see~\ref{sec:dataset}). This focused yet heterogeneous domain enables rigorous testing of LAVA’s ability to produce accurate outputs and avoid false positives, particularly where heuristic, zero-shot, and few-shot LLM-based methods struggle with layout variability or reasoning complexity.

To ensure rigorous and interpretable evaluation, we adopt rule-level testing rather than intent-level, with validation rules drawn from a curated library provided by business stakeholders. Each rule corresponds to an atomic validation unit and is applied only to the document types for which it is defined. For fairness and comparability, document type metadata is included with the dataset, since accurate validation cannot be assessed without first matching each document to its correct rule set; this ensures that all pipelines are evaluated under the same conditions, without document and validation retrieval. Ground-truth outcomes are manually annotated by domain experts, and automatic evaluation is complemented by manual audits from business collaborators, ensuring both accuracy and institutional credibility in line with production-grade expectations for document validation.

\subsection{Implementations}

To ensure fairness, all baselines, LAVA, and LAVA's ablation variants use Claude 3.7 Sonnet~\cite{claude37} as the validation component. We configure the model with a maximum response length of 5120 tokens and enable reasoning (thinking) with a budget of 1024 tokens, allowing the model to better understand complex validation rules and generalize across documents with diverse layouts and edge scenarios. For LAVA’s Arithmetic Processor, the maximum retry
number $n$ in the fallback loop is set to $2$, to ensure a practical trade-off between error correction and computational efficiency. And all other LLM parameters are kept identical across evaluations.

For document analysis in experiments, we prioritized a methodology that ensures our findings are portable, reproducible, and immediately accessible. To this end, our experiment exclusively utilizes off-the-shelf tools that do not require custom training or dataset-specific fine-tuning. A foundational layer of raw text and spatial information was established for the experiment using the open-source Tesseract OCR~\cite{TessOverview}. For the processing of structural semantics, such as tables and forms, the publicly available AWS Textract service was employed. This consistent and transparent tooling strategy not only ensures that the comparative evaluation in Sections~\ref{sec:baselines} and \ref{sec:ablation} fairly assesses the core architectural and reasoning capabilities of the different pipelines, but also directly supports other researchers to easily replicate and build upon our results.

\subsection{Validation Rules} \label{sec:rules}
We group several dozen validation rules into five categories, reflecting distinct reasoning and computation demands. These categories assess the pipeline’s ability to extract information, integrate rules, and perform contextual reasoning. See~\ref{sec:appendix_rules} for complete versions of some representative rules.







\subsection{Evaluation Metrics} \label{sec:metrics}
Our evaluation metrics comprehensively assess reasoning quality across all pipeline stages. Given the distinct characteristics of our compliance-sensitive validation task compared to common document understanding settings, we design task-specific metrics with particular emphasis on suppressing false positives---a critical industrial concern causing costly investigations, delays, and loss of trust. We evaluate baselines from three perspectives to cover these aspects. Their formal definitions are given in Appendix~\ref{app:metrics}, along with visual examples showing how each metric manifests in an illustrative sample document.

\paragraph{Hallucination Control.}
We report the percentage of responses with two failure types:
\begin{itemize}
    \item \textbf{Factual Hallucination Rate:} Content not grounded in the source document, such as fabricated values or formulas, and unsupported claims.
    \item \textbf{Numerical Infidelity Rate:} Incorrect quantitative reasoning or numerical derivations, such as incorrect formula execution results and conclusions inconsistent with the given numerical evidence.
\end{itemize}

\paragraph{Edge Case Handling.}
This metric measures failure rate on complex or exception-driven scenarios, about 10\%–25\% of document–rule pairs. Such cases require \textit{adaptive logic reasoning} beyond fixed rules and broader \textit{logic coverage} for diverse edge conditions. 
As these challenges often overlap, we report a single rate to capture both, where higher values indicate weaker generalization and reasoning flexibility.


\paragraph{Token Cost.} We measure input and output tokens per rule check to assess computational efficiency and deployment cost, highlighting trade-offs between reasoning quality and efficiency across pipelines. For each document-rule pair, token counts are recorded and aggregated over the full evaluation set.


\begin{table*}[t]
\centering
\setlength{\tabcolsep}{5pt}
\newcolumntype{C}[1]{>{\centering\arraybackslash}p{#1}}
\small
\begin{tabular}{p{1.7cm}|C{0.9cm}|C{3.4cm}|C{3.4cm}|C{3.2cm}|C{1.2cm}}
\toprule
\textbf{Metric} & \textbf{Rule Group} & \textbf{VLM + Field-Level OCR} & \textbf{LLM + Field-Level OCR} & \textbf{LLM + Enhanced OCR} & \textbf{LAVA} \\
\midrule
Factual         & $C_1$ & 0.03 & 0.03 & 0.01 & \textbf{0.01} \\
Hallucination   & $C_2$ & 0.08 & 0.10 & 0.04 & \textbf{0.03} \\
Rate            & $C_3$ & 0.31 & 0.33 & 0.30 & \textbf{0.18} \\
                & $C_4$ & 0.05 & 0.08 & 0.05 & \textbf{0.03} \\
                & $C_5$ & 0.28 & 0.30 & 0.15 & \textbf{0.05} \\
\midrule
Numerical        & $C_1$ & N/A & N/A & N/A & \textbf{N/A} \\
Infidelity       & $C_2$ & 0.08 & 0.04 & 0.03 & \textbf{0.01} \\
Rate             & $C_3$ & 0.33 & 0.31 & 0.20 & \textbf{0.10} \\
                 & $C_4$ & 0.11 & 0.08 & 0.05 & \textbf{0.00} \\
                 & $C_5$ & 0.25 & 0.10 & 0.10 & \textbf{0.00} \\
\midrule
Edge Case        & $C_1$ & N/A & N/A & N/A & \textbf{N/A} \\
Error Rate       & $C_2$ & 0.27 & 0.25 & 0.23 & \textbf{0.02} \\
                 & $C_3$ & 0.86 & 0.90 & 0.82 & \textbf{0.17} \\
                 & $C_4$ & 0.46  & 0.34  & 0.43  & \textbf{0.11} \\
                 & $C_5$ & 0.89  & 0.92  & 0.86  & \textbf{0.08} \\
\bottomrule
\end{tabular}
\vspace{-1em}
\caption{Performance of baselines across three metrics and five validation rule categories ($C_1$–$C_5$). $C_1$: content extraction; $C_2$: conditional logic reasoning; $C_3$: multi-step logic reasoning; $C_4$: unconstrained arithmetic consistency checking; $C_5$: constrained arithmetic consistency checking.
}

\label{table:pipelines_rate}
\end{table*}

\subsection{Comparative Baselines}\label{sec:baselines}
Here, \textit{Field-Level OCR} denotes OCR output containing only individual field texts and their spatial information (e.g., from bounding boxes), without higher-level structure such as tables, sections, or relational links between fields. This representation preserves raw layout signals but omits the structured markup used in LAVA’s pipeline. 

We consider three representative settings:
\begin{itemize}
    \item \textbf{VLM + Field-Level OCR:} 
    Document images with Field-Level OCR in a VQA setup, where OCR text aids visual grounding.
    
    \item \textbf{LLM + Field-Level OCR:} 
    Same OCR content without images, measuring performance from textual–spatial data alone in a QA setup.
    
    \item \textbf{LLM + Enhanced OCR (no structural markup):} 
    OCR refined with domain-specific cleanup and recovery steps identical to LAVA’s extraction, but without structured markup, isolating LAVA's impact of structured representation and modular reasoning.
\end{itemize}

Since OCR and markup information might introduce noise (e.g., when text is misrecognized), we also examine an OCR-free configuration to assess the impact of textual knowledge. This setting is evaluated in ablation study (Section \ref{sec:ablation}), where all other modules are preserved but the model operates only on document images. It is excluded from the baseline set because the baselines are intended to preserve textual content in some form for fair comparison, whereas this variant probes the limits of visual-only reasoning and represents a more extreme ablation of the pipeline.

As shown in Table~\ref{table:pipelines_rate}, LAVA achieves the lowest failure rates across all metrics and categories, with especially large gains in multi-step logic reasoning ($C_3$) and constrained arithmetic consistency checking ($C_5$), reducing hallucination and numerical errors by over 10\% compared to the best baseline. And it records approximately 0\% Numerical Infidelity Rate in three of four categories. LAVA also substantially lowers edge case errors, indicating strong generalization under layout ambiguity and rare conditions. Common VQA and QA settings represented by the baselines underperform in this compliance-oriented validation task due to their limited handling of layout variability and domain-specific rules. These results highlight the benefit of structured prompts and modular reasoning in suppressing unsupported content generation and enabling reliable multi-step, context-aware reasoning over structural data.

We also sampled a subset of DocVQA~\cite{mathew2021docvqadatasetvqadocument} 
documents with five manually defined rules each, where LAVA also achieved strong performance, confirming that our framework readily transfers to public data. More importantly, real-world validation tasks---where both documents and rules are richer in semantics, context, and reasoning complexity---pose substantially 
greater challenges, where our approach proves particularly effective under industry-level conditions.


\subsection{Ablation Study}\label{sec:ablation}
To assess the contribution of each core module in LAVA, we performed an ablation study using a unified metric: the percentage of responses that fail to exactly match the ground truth. This stricter measure was chosen because LAVA’s earlier evaluations achieved low error rates in category-specific metrics, so a single comprehensive indicator better reveals performance drops when modules are altered or removed.

We evaluated six configurations: the full LAVA system; three \textbf{Knowledge Extraction (KE)} variants; and versions without \textbf{Info Augmentation (IA)} or \textbf{Arithmetic Processor (AP)}. The KE variants progressively reduce structural detail and change input format:
\begin{enumerate}
    \item \textbf{Markdown KE} replaces LAVA’s structured markup with Markdown-like fashion.
    \item \textbf{Plain-text KE} flattens layout into text aligned to mimic visual spacing but without structural tags.
    \item \textbf{No KE} omits textual knowledge entirely, prompting the model with original document images. This configuration serves as the OCR-free, vision-only benchmark we referred to in the baseline comparison (Section~\ref{sec:baselines}).
\end{enumerate}

This stepwise removal strips explicit layout and relational cues, forcing reliance on surface text or raw images. For AP ablation, AP is replaced by Symbolic Reasoner, removing explicit calculation/result validation. The two Retrieval modules are not ablated since all experiments use a fixed rule set, making them independent of validation accuracy.

\begin{table}[t]
\centering
\setlength{\tabcolsep}{4pt}
\newcolumntype{C}[1]{>{\centering\arraybackslash}p{#1}}
\small
\begin{tabular}{l|*{5}{C{0.55cm}}}
\toprule
\textbf{Pipeline} & $C_1$ & $C_2$ & $C_3$ & $C_4$ & $C_5$ \\
\midrule
LAVA w/o KE & 0.03 & 0.65 & 0.67 & 0.09 & 0.66 \\
LAVA w/ Plain-text KE & 0.02 & 0.63 & 0.58 & 0.08 & 0.48 \\
LAVA w/ Markdown KE & 0.01 & 0.25 & 0.54 & 0.05 & 0.55 \\
LAVA w/o IA & 0.01 & 0.15 & 0.45 & 0.05 & 0.12 \\
LAVA w/o AP & 0.01 & 0.03 & 0.29 & 0.10 & 0.56 \\
\textbf{LAVA} & \textbf{0.01} & \textbf{0.03} & \textbf{0.28} & \textbf{0.05} & \textbf{0.07} \\
\bottomrule
\end{tabular}
\vspace{-1em}
\caption{
Percentage of responses failing to exactly match ground-truth answers across five rule categories.
}
\label{table:ablation}
\end{table}

The error patterns in table \ref{table:ablation} shows that KE, IA, and AP play complementary roles, with KE being the most critical. Removing KE forces the model to rely almost entirely on visual reasoning, raising $C_2$, $C_3$, and $C_5$ failure rate to 0.65–0.67—over twice LAVA's error in multi-step logic ($C_3$) and nearly tenfold in constrained arithmetic consistency checking ($C_5$) even with the validation module's assistance.
Two factors explain this:
(1) Structural conversion: without KE’s structural format that preserves complex and nested layout, grouping, and field dependencies, multi-step logic and table reasoning lose explicit relational cues. Downgrading to Markdown or plain text progressively strips away these signals, with Markdown underperforming plain text in $C_5$ due to weaker preservation of columnar alignment in tables;
(2) Content filtering: without KE’s filtering stage, noisy headers, footers, and irrelevant fields dilute attention and amplify hallucination risk.
This gradient from fully structured to none shows that noise suppression, explicit spatial and relational cues are essential for stable reasoning, especially in multi-field and arithmetic-heavy rules for financial document validation tasks.

IA mainly impacts logical reasoning: removing it raises $C_3$ error from 0.28 to 0.45, suggesting that metadata and domain cues help models resolve field semantics and linking conditions across steps.

AP’s effect is computation-focused: removing it leaves $C_1$–$C_3$ unchanged but drives $C_4$/$C_5$ error to 0.10/0.56, reflecting the limits of unconstrained LLM reasoning in arithmetic tasks and the benefit of explicit calculation with fallback mechanism.

Overall, the degradations are consistent with the design intent: KE supplies the structured substrate, IA enriches it with semantic context, and AP enforces numerical fidelity, justifying LAVA’s functionally complementary and modular composition.

\subsection{Business Impact}

\begin{figure}[t]
    \centering
    \begin{subfigure}[t]{0.48\textwidth}
        \centering
        \includegraphics[width=\linewidth]{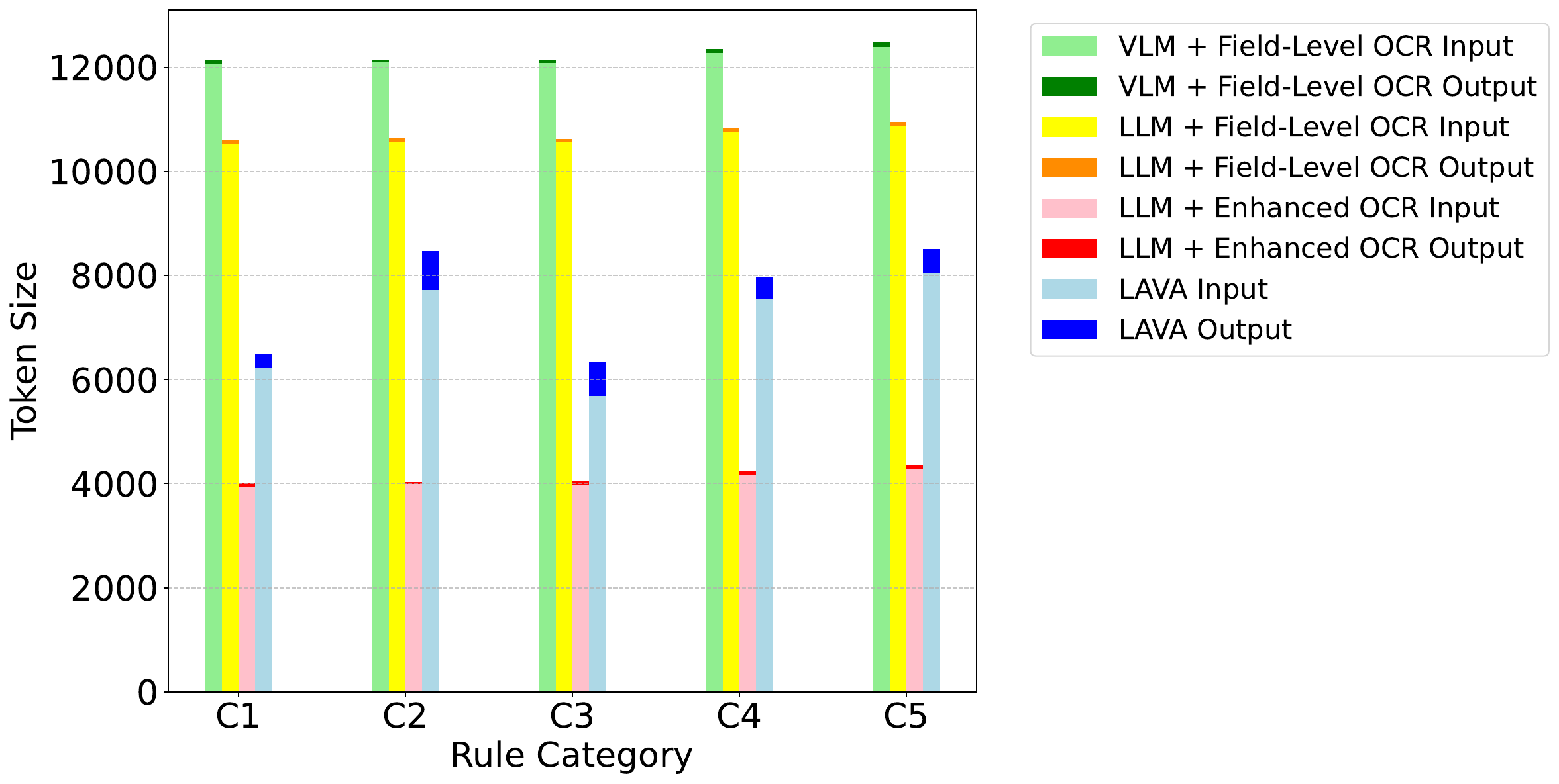}
        \caption{LAVA and baselines.}
        \label{fig:pipelines_token}
    \end{subfigure}
    \hfill
    \begin{subfigure}[t]{0.48\textwidth}
        \centering
        \includegraphics[width=\linewidth]{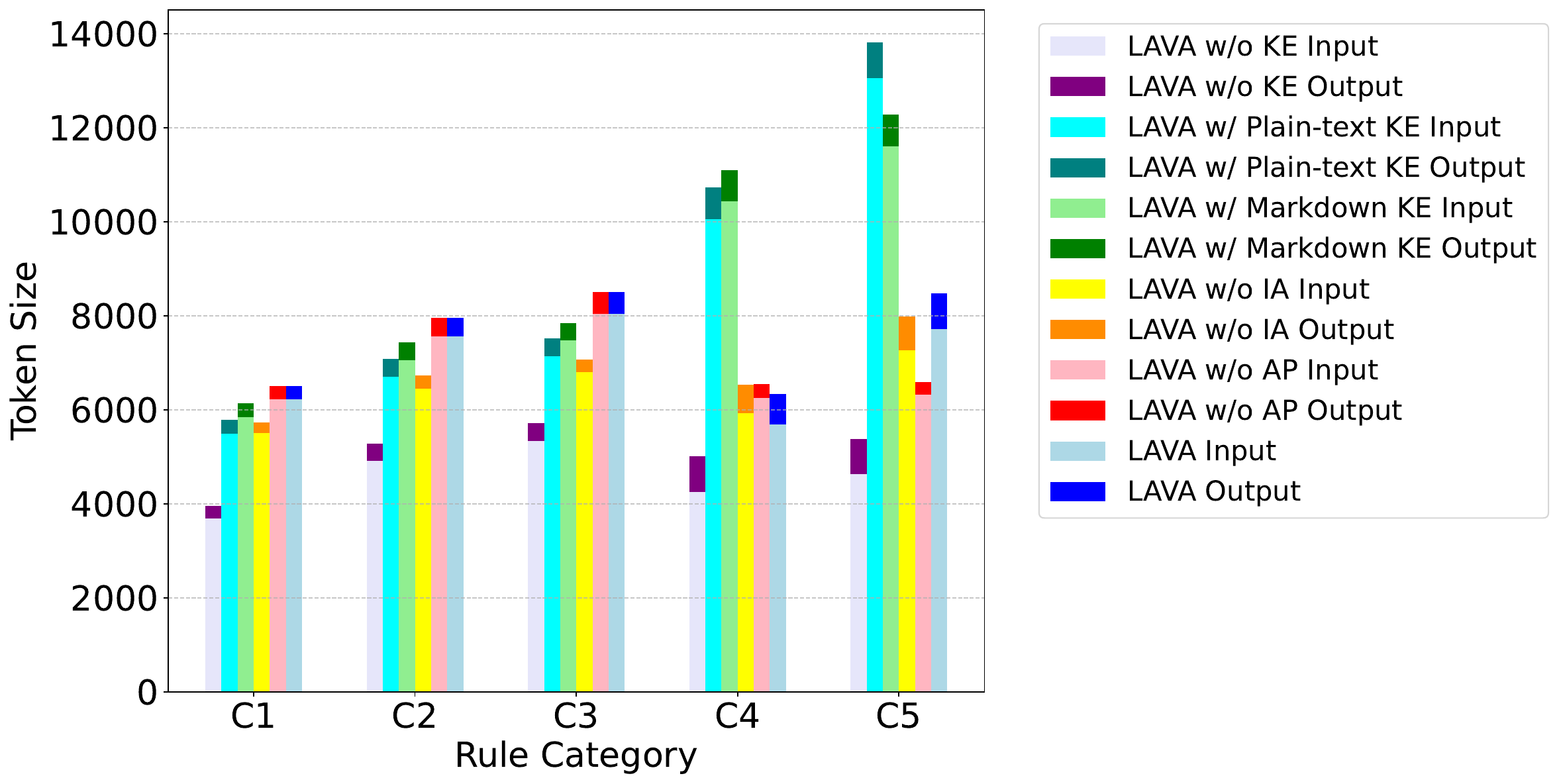}
        \caption{LAVA and ablated variants.}
        \label{fig:ablation_tokens}
    \end{subfigure}
    \vspace{-1em}
    \caption{Average input and output token counts across five rule categories.}
    \label{fig:token_counts}
\end{figure}

While Section~\ref{sec:baselines} and~\ref{sec:ablation} highlight LAVA’s technical advantages in accuracy and robustness, real-world adoption also depends on its operational efficiency and cost-effectiveness. By consolidating token usage, we can further assess how LAVA scales under realistic operational constraints, using token consumption as a proxy for latency and API expenditure in high-volume production workflows.

LAVA amortizes multi-step costs by extracting structured knowledge once and reusing it across rules. As shown in Figure~\ref{fig:pipelines_token}, this design cuts input tokens by 25\%–45\% versus VLM and LLM baselines with Field-Level OCR, lowering inference costs and enabling faster turnaround for time-critical financial validation processes, while also achieving fewer errors (Table~\ref{table:pipelines_rate}). 

Ablation results (Figure~\ref{fig:ablation_tokens}) show that in non-computation tasks ($C_1$–$C_3$), excluding variant w/o KE, LAVA adds minimal tokens with stable output length due to the predefined schema-based response format. For computation tasks ($C_4$–$C_5$), the less faithful structural representation from plain-text and Markdown KE inflate inputs by causing more model output errors and retries in the validation loop. Across all ablations, full LAVA adds under 3k tokens per rule (\(\sim\$0.009\) with Claude 3.7 or \(\sim\$0.006\) with GPT-4.1), an overhead outweighed by substantial gains in accuracy and reasoning stability. This demonstrates that LAVA delivers a dual advantage: (1) immediate computational savings from reduced token usage, and (2) long-term operational efficiency from its modular, auditable design. This combination suggests that LAVA's architectural benefits outweigh the modular integration complexity, making it a cost-effective and scalable choice for high-volume, time-critical real-world financial workflows.

\section{Conclusion}

We presented LAVA, a modular, interpretable and backbone- and domain-agnostic system for enterprise-grade financial document validation. Combining layout-preserving extraction, domain-aware augmentation, and symbolic and arithmetic verification, LAVA achieves high factual alignment and numerical reliability on a large real-world benchmark, with lower computational overhead than monolithic prompting. Evaluation with factual hallucination rate, numerical infidelity rate, and edge case handling shows that structured, multi-stage reasoning enables fine-grained error attribution, stronger robustness in compliance-sensitive workflows, and transfer to structurally similar data. Looking ahead, we aim to handle noisier, less-structured documents and to learn rule generalization across formats and domains.

\section{Ethical Considerations}

This research was conducted using a proprietary dataset of financial documents, with all data fully anonymized and handled under strict institutional privacy protocols. We acknowledge the potential risk of algorithmic bias in financial decision-making. LAVA's design as an accurate, scalable and auditable system is a deliberate step to mitigate this risk by enhancing transparency and ensuring human accountability. Nevertheless, any production deployment would necessitate rigorous, ongoing audits for demographic bias to ensure fair and equitable outcomes.

\clearpage
\phantomsection
\bibliography{custom}

\clearpage
\appendix
\twocolumn[
\section{Appendix}\label{sec:appendix}
]
\setlist[enumerate]{leftmargin=1.5em, itemsep=0pt, topsep=0pt}
\setlist[itemize]{
  leftmargin=1em,  
  itemsep=0pt,     
  topsep=-3pt       
}

\newcommand{\variable}[1]{\textcolor{blue}{\texttt{\{#1\}}}}

\tcbset{
  prompt/.style={
    width=\linewidth,
    colframe=Periwinkle,            
    colback=Lavender!10,            
    colbacktitle=Periwinkle,        
    coltitle=white,                 
    boxrule=0.5pt,
    sharp corners=south,
    fonttitle=\bfseries,
    before upper={\scriptsize\setstretch{0.9}\setlength{\parindent}{0pt}},
    breakable
  }
}

\tcbset{
  output/.style={
    width=\linewidth,
    colframe=RoyalBlue,             
    colback=SkyBlue!10,             
    colbacktitle=RoyalBlue!80,      
    coltitle=white,                 
    boxrule=0.5pt,
    sharp corners=south,
    fonttitle=\bfseries,
    title={\small\textbf{Output Example}},
    before upper={\scriptsize\setstretch{0.9}\setlength{\parindent}{0pt}},
    breakable
  }
}

\tcbset{
  ground_truth/.style={
    width=\linewidth,
    colframe=ForestGreen,           
    colback=Green!10,               
    colbacktitle=ForestGreen!80,    
    coltitle=white,                 
    boxrule=0.5pt,
    sharp corners=south,
    fonttitle=\bfseries,
    title={\small\textbf{Ground Truth Output}},
    before upper={\scriptsize\setstretch{0.9}\setlength{\parindent}{0pt}},
    breakable
  }
}

\tcbset{
  error/.style={
    width=\linewidth,
    colframe=BrickRed,           
    colback=Red!10,              
    colbacktitle=BrickRed!80,    
    coltitle=white,              
    boxrule=0.5pt,
    sharp corners=south,
    fonttitle=\bfseries,
    before upper={\scriptsize\setstretch{0.9}\setlength{\parindent}{0pt}},
    breakable
  }
}

\subsection{Prompt used in LAVA}
\subsubsection{Prompt for Symbolic Reasoner}
\label{sec:prompt1}

\begin{tcolorbox}[prompt, title={\small\textbf{Dynamically Generated Prompt}}]
You are an assistant who verifies financial documents. You are given\\ \variable{document\_types} (in \variable{languages}) and extracted\\\variable{knowledge\_names}. Your job is to verify the documents based on the validation rule. Please do not make up any values.

\vspace{1em}
<instructions>
\begin{enumerate}
    \item Use structured representation as references for identifying the layout and structure of the documents.\\
    <html\_tables>\variable{knowledge[tables]}</html\_tables>\\
    <forms>\variable{knowledge[forms]}</forms>\\
    ...
    \item Use the semantic fields as references for providing explicit contextual information.\\
    <semantic\_fields>\variable{knowledge[fields]}</semantic\_fields>
    \item[...]
    \item[\textit{n}.] Output as follows:
    \begin{itemize}
        \item Only output the fields requested in the validation rule, do not include any other fields stated in example\_output.
        \item Replace all spaces in keys with underscores "\_".
        \item If any field is not present, set value to "".
        \item  Output date in YYYYMMDD format.
    \end{itemize}
    <example\_output>\variable{example\_output}</example\_output>
\end{enumerate}
</instructions>

\vspace{1em}
<validation\_rule>\\
\variable{validation\_rule}
\begin{itemize}[leftmargin=1.5em, itemsep=-3pt, topsep=0pt]
    \item \variable{domain\_specific\_term\_1} : \variable{explanation\_1}
    \item[...]
    \item \variable{domain\_specific\_term\_n} : \variable{explanation\_n}
\end{itemize}
</validation\_rule>

\vspace{1em}
<output\_format>\\
Return solely the JSON object without any additional explanations, comments, preamble, or formatting like backticks.\\
</output\_format>
\end{tcolorbox}

\subsubsection{Prompt for Arithmetic Processor}
\label{sec:prompt2}

\begin{tcolorbox}[prompt, title={\small\textbf{Prompt for Formula Generation}}]
\textbf{System Prompt:}\\
You are an assistant who generates a Python-style calculation formula to explain how a request field should be computed.
You are given \variable{document\_types} (in \variable{languages}), as well as knowledge extracted from the documents and a validation rule.

\vspace{1em}
<instructions>
\begin{enumerate}
    \item Identify the requested field from the rule. Store its name in the JSON object with the key "field\_name", delete any special symbols and replace all spaces in keys with underscores "\_".
    \item Extract the stated value for the requested field.
    \begin{itemize}
        \item Store it in the JSON object with the key "stated" and make sure it is a valid float or integer. 
        \item If the requested field is not present or empty, set "stated" to "NaN".
        \item Do not make up any values.
    \end{itemize}
    \item Identify the relevant numerical fields in the knowledge that are needed to compute the requested field. Consider how those fields logically combine to produce the value of the requested field.
    \item Once you find an appropriate calculation expression, store it in Python-executable format in the JSON object with the key "formula":
    \begin{itemize}
        \item Do not compute the result.
        \item Do not include any functions like "round(...)" or similar.
        \item Only use raw numerical operations.
    \end{itemize}
    \item Return solely the JSON object without any additional explanations, comments, preamble, or formatting like backticks.
\end{enumerate}
</instructions>

\vspace{1em}
<example\_output>\\
\{\\
\hspace*{1.5em} "field\_name": "Current\_Total\_Income",\\
\hspace*{1.5em} "stated": 600,\\
\hspace*{1.5em} "formula": "100 + 200 + 300"\\
\}\\
</example\_output>

\vspace{2em}
\textbf{User Prompt:}\\
<knowledge>\variable{knowledge}</knowledge>

\vspace{1em}
<validation\_rule>\\
\variable{validation\_rule}
\begin{itemize}[leftmargin=1.5em, itemsep=-3pt, topsep=0pt]
    \item \variable{domain\_specific\_term\_1} : \variable{explanation\_1}
    \item[...]
    \item \variable{domain\_specific\_term\_n} : \variable{explanation\_n}
\end{itemize}
</validation\_rule>
\end{tcolorbox}

\begin{tcolorbox}[prompt, title={\small\textbf{Prompt for Formula Correction}}]

\textbf{System Prompt:}\\
You are the greatest financial auditor, logician and deducer.
You are given \variable{document\_types} (in \variable{languages}), as well as knowledge extracted from the documents, a validation rule, and a calculation expression.

\vspace{1em}
<instructions>
\begin{enumerate}
    \item The calculation expression in response to the validation rule is wrong based on the knowledge extracted.
    \item Check if there is any other correct calculation expression that can be derived from the data given. 
    \item Once you find an appropriate calculation expression, store it in Python-executable format:
        \begin{itemize}
            \item Do not compute the result using equal sign.
            \item Do not include any functions like "round(...)" or similar.
            \item Only use raw numerical operations.
        \end{itemize}
    \item Return solely the calculation expression without any additional explanations, comments, preamble, or formatting like backticks.
\end{enumerate}
</instructions>

\vspace{1em}
<example\_output\_1>100 + 200 + 300</example\_output\_1>\\
<example\_output\_2>80 * 100.5</example\_output\_2>

\vspace{2em}
\textbf{User Prompt:}\\
<knowledge>\variable{knowledge}</knowledge>

\vspace{1em}
<wrong\_calculation>\variable{wrong\_calculation}<wrong\_calculation>

\vspace{1em}
<validation\_rule>\\
\variable{validation\_rule}
\begin{itemize}[leftmargin=1.5em, itemsep=-3pt, topsep=0pt]
    \item \variable{domain\_specific\_term\_1} : \variable{explanation\_1}
    \item[...]
    \item \variable{domain\_specific\_term\_n} : \variable{explanation\_n}
\end{itemize}
</validation\_rule>
\end{tcolorbox}

\subsection{Representative Validation Rules and Output Examples Used in Experiment} \label{sec:appendix_rules}
\begin{enumerate}
    \item Content Extraction\\
    \textbf{Rule:}
    \begin{enumerate}
        \item Are Current and YTD regular pay/salary amounts present in documents?
        \item Is Current and YTD CPP/QPP (may appear as Government Pension) present in the paystub?
        \item Is Social Insurance Number (SIN) present in the T4 document?
    \end{enumerate}
    \begin{tcolorbox}[output]
        \{\\
            \hspace*{1.5em} "Employer\_Name": \{"present": true, "value": "organization"\},\\
            \hspace*{1.5em} "Current\_Regular\_Pay": \{"present": false, "value": ""\},\\
        \}
    \end{tcolorbox}
    \item Conditional Logic Reasoning
    \begin{enumerate}
        \item \textbf{Rule:}\\
            <CRA\_EI\_data>\variable{EI\_data}</CRA\_EI\_data>\\
            <EI\_verification\_rules>
            \begin{itemize}
                \item When Current EI is non-zero, skip the check and set "valid" to true.
                \item When Current and YTD EI are both 0 or blank, skip the check and set "valid" to false. 
                \item When Current EI is 0 or blank, YTD EI is not 0 or blank. If YTD EI $\leq$ EI cap, set "valid" to true. Otherwise, set "valid" to false. 
            \end{itemize}
            </EI\_verification\_rules>\\
            Do the EI values comply with rules based on the Canada Revenue Agency (CRA) guidelines?
            \begin{tcolorbox}[output]
                \{\\
                    \hspace*{1.5em} "Pay\_Date": "20241128",\\
                    \hspace*{1.5em} "EI": \{"Current": "100", "YTD": "200", "Cap": "1049.12", "valid": true\}\\
                \}
            \end{tcolorbox}
        \item \textbf{Rule:}\\
            <numerical\_verification\_rules>
            \begin{itemize}
                \item Cells with empty value, or only numbers, symbols, punctuation are valid.
                \item Cells containing words (alphabetic characters) are invalid.
                \item If all fields are valid, return an empty object \{\}.
            \end{itemize}
            <numerical\_verification\_rules>\\
            Based on the rules provided, does every non-header cell in tables contain a valid numerical value?
            \begin{tcolorbox}[output]
                \{\\
                    \hspace*{1.5em} "This\_Period\_Regular": "amount",\\
                    \hspace*{1.5em} "YTD\_CPP": "value"\\
                \}
            \end{tcolorbox}
    \end{enumerate}
    \item Multi-step Logic Reasoning\\
        \textbf{Rule:}\\
        <pay\_freq\_cap>\\
            $[$\\
                \hspace*{1.5em} \{"freq": "monthly", "cap": 12\},\\
                \hspace*{1.5em} \{"freq": "semimonthly", "cap": 24\},\\
                \hspace*{1.5em} \{"freq": "biweekly", "cap": 26\},\\
                \hspace*{1.5em} \{"freq": "weekly", "cap": 52\}\\
            $]$\\
        </pay\_freq\_cap>\\
        <freq\_verification\_rules>
        \begin{itemize}
            \item If start or end date is empty, set "stated" to "" and set "equal" to None.
            \item If No. of pay period is empty, set "stated" to "" and set "equal" to None.
            \item If No. of pay period is not empty, find its corresponding "frequency cap" from pay frequency rules:
                \begin{itemize}
                    \item Compute the expected "current period number": Use "Pay Frequency", "Start Date" and "frequency cap" to calculate how many periods have elapsed since the start of the year.
                    \item Extract current period number from No. of pay period stated (e.g., 22 on "22 of 26").
                    \item Check if the "stated" current period number" equals the "expected current period number". If both equal, set "equal" to true. Otherwise, set "equal" to false.
                \end{itemize}
        \end{itemize}
        </freq\_verification\_rules>\\
        Based on the rules provided, does pay frequency and No. of pay period in the extracted date comply with the pay frequency rules?
        \begin{tcolorbox}[output]
            \{\\
                \hspace*{1.5em} "Start\_Date": "20240101",\\
                \hspace*{1.5em} "End\_Date": "20240131",\\
                \hspace*{1.5em} "Pay\_Frequency": \{"stated": "semimonthly", "expected": "monthly", "equal": false\},\\
                \hspace*{1.5em} "NO\_Pay\_Period": \{"stated": "11", "expected": "1", "equal": false\}\\
            \}
        \end{tcolorbox}
    \item Unconstrained Arithmetic Consistency Checking\\
    \textbf{Rule:}
    \begin{enumerate}
        \item Is the YTD gross pay calculated correctly by summing up all earning items?
        \item Is Line 23600 (Net Income) correctly calculated as Line 15000 (Total Income) minus Deductions from Total Income?
        \item Is the ending balance correctly calculated as the starting balance plus total deposits minus total withdrawals for the period?
    \end{enumerate}
    \begin{tcolorbox}[output]
        \{\\
            \hspace*{1.5em} "Current\_Gross\_Pay": \{"calculation": "100 + 200 + 300 = 600", "stated": "300", "equal": true\},\\
            \hspace*{1.5em} "YTD\_Gross\_Pay": \{"calculation": "1500 + 200 + 300 = 2000", "stated": "", "equal": null\}\\
        \}
    \end{tcolorbox}
    
    \item Constrained Arithmetic Consistency Checking\\
    \textbf{Rule:}
    \begin{enumerate}
        \item Is the Current regular pay calculated correctly by rate * hours when rounding the result to 2 decimal places? Set formula to an empty string"" when rate or hours is not present or 0.
        \item Is the current net pay correctly computed as the current gross pay minus all current deductions, with the result rounded to 2 decimal places? If the table includes a "Total Deduction" field, use it directly instead of summing individual deduction items.
    \end{enumerate}
    \begin{tcolorbox}[output]
        \{\\
            \hspace*{1.5em} "Current\_Gross\_Pay": \{"calculation": "100 + 200 + 300 = 600", "stated": "300", "equal": true\},\\
            \hspace*{1.5em} "YTD\_Gross\_Pay": \{"calculation": "1500 + 200 + 300 = 2000", "stated": "", "equal": null\}\\
        \}
    \end{tcolorbox}
\end{enumerate}

\subsection{Evaluation Metrics} \label{app:metrics}

\subsubsection{Factual Hallucination Rate (FHR).}\label{app:metrics_FHR}
Given a fixed rule $r$ with evaluation document set 
$\mathcal{D}_r=\{d_1,\dots,d_{n_r}\}$,
let $\hat{\mathcal{E}}_{i,r}$ denote the predicted evidence subset in the model’s output explanation trace for $(d_i,r)$,
and $\mathcal{E}(d_i,r)$ the gold evidence set from $d_i$.
Let $\hat{\varphi}_{i,r}$ denote the predicted formula for $(d_i,r)$,
and $g_r$ the corresponding gold formula.

We mark a sample $(d_i,r)$ as factually hallucinated if either hold:
\begin{enumerate}[label=(\alph*),leftmargin=1em]
\item \textbf{Evidence Hallucination:} 
      $\hat{\mathcal{E}}_{i,r}\not\subseteq \mathcal{E}(d_i,r)$.
\item \textbf{Formula Hallucination:} 
      $\hat{\varphi}_{i,r}\not\equiv g_r$ 
      (tested via polynomial identity testing with Schwartz–Zippel).
\end{enumerate}

Formally, the sample-level indicator is
\begin{equation}
\label{eq:fhr-sample}
\begin{aligned}
\mathrm{FH}_{i,r} = \mathbf{1}\!\big[ \;&
  (\hat{\mathcal{E}}_{i,r}\not\subseteq \mathcal{E}(d_i,r)) \;\;\vee \\ 
  &(\hat{\varphi}_{i,r}\not\equiv g_r) \big].
\end{aligned}
\end{equation}

The rule-level rate is
\begin{equation}
\label{eq:fhr-rule}
\mathrm{FHR}_{r}
= \frac{1}{|\mathcal{D}_r|}
  \sum_{d_i\in\mathcal{D}_r}\mathrm{FH}_{i,r},
\end{equation}
and the group-level average for $C\subseteq\mathcal{R}$ is
\begin{equation}
\label{eq:fhr-group-w}
\mathrm{FHR}^{\mathrm{w}}_{C}
= \frac{\sum_{r\in C}|\mathcal{D}_r|\cdot \mathrm{FHR}_{r}}
       {\sum_{r\in C}|\mathcal{D}_r|}.
\end{equation}

Containment checks normalize case, punctuation, numeric formatting, and unit conventions. 
Formula equivalence ($\equiv$) is evaluated by polynomial identity testing under randomized substitution.

\begin{figure}[t]
  \centering
  \includegraphics[width=\linewidth]{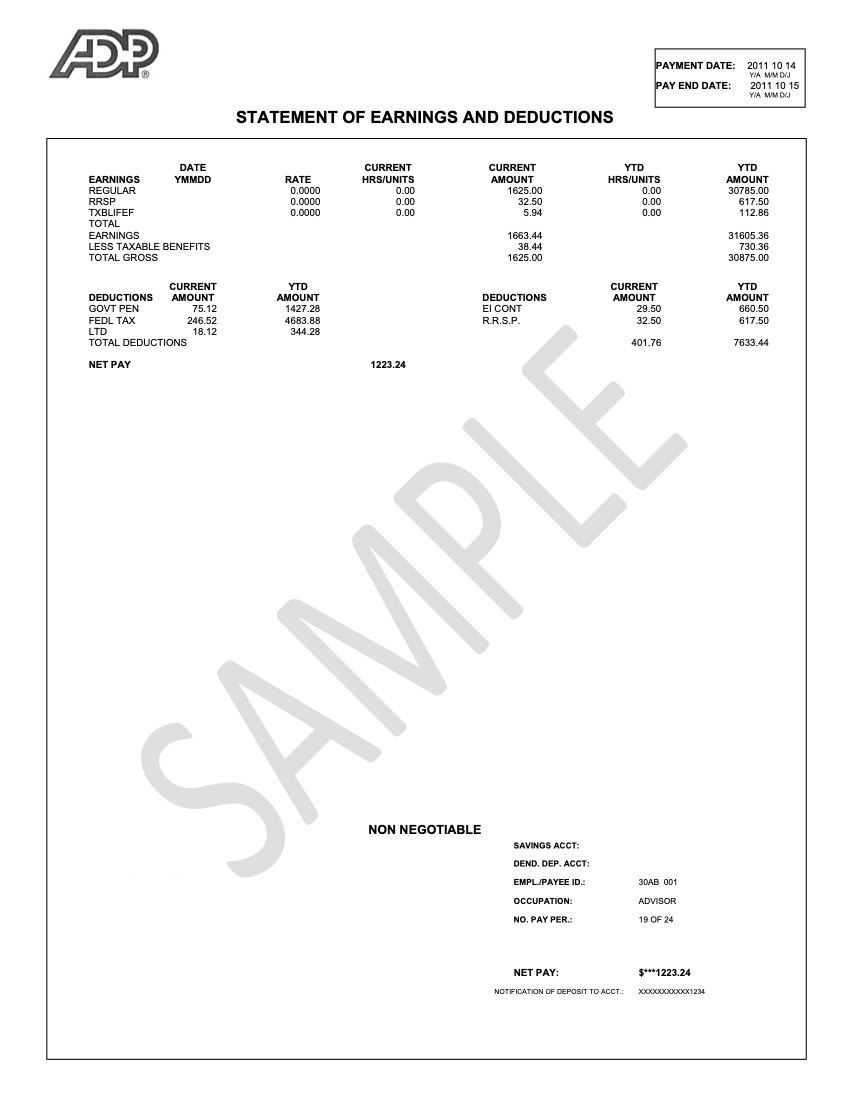}
  \caption{Sample document shown for clarity and error illustrations; metrics are document-agnostic and applied unchanged across all document types (sensitive content redacted).}
  \label{fig:paystub_sample}
\end{figure}

\paragraph{Example.}  
Consider the rule “Is the YTD total deduction computed as the sum of all YTD items in the deduction table?”
The pipeline is asked to extract relevant evidence, provide a calculation, and conclude the validation result for this task.
Based on Figure~\ref{fig:paystub_sample}, the ground truth is:
\begin{tcolorbox}[ground_truth]
    \{\\
        \hspace*{1.5em} "YTD\_Total\_Deduction": \{\\
        \hspace*{3em}"calculation": "1427.28 + 4683.88 + 344.28 + \textbf{660.50} + 617.50 = 7733.44",\\
        \hspace*{3em}"stated": "7633.44",\\
        \hspace*{3em}"equal": false\\
        \hspace*{1.5em}\}\\
    \}
\end{tcolorbox}
When \emph{factual hallucination} occurs, the pipeline fabricates one or some of the variables in the calculation formula (e.g., changing $660.50$ to $560.50$) so that the sum exactly matches the stated total:
\begin{tcolorbox}[error, title={\small\textbf{Output with Factual Hallucination}}]
    \{\\
        \hspace*{1.5em} "YTD\_Total\_Deduction": \{\\
        \hspace*{3em}"calculation": "1427.28 + 4683.88 + 344.28 + \textbf{560.50} + 617.50 = 7633.44",\\
        \hspace*{3em}"stated": "7633.44",\\
        \hspace*{3em}"equal": true\\
        \hspace*{1.5em}\},\\
    \}
\end{tcolorbox}
Consider the rule: “Are pay period start and end dates present in the document?” 
The pipeline is asked to check the presence of specific fields and extract the corresponding value.
The ground truth for Figure~\ref{fig:paystub_sample} is:
\begin{tcolorbox}[ground_truth]
    \{\\
        \hspace*{1.5em} "Start\_Date": \{"present": \textbf{false}, "value": \textbf{""}\}\\
        \hspace*{1.5em} "End\_Date": \{"present": true, "value": "20111015"\}\\
    \}
\end{tcolorbox}
When \emph{factual hallucination} occurs, the pipeline fabricates a start date by using the value of another field (in this case, the payment date) and incorrectly marking it as present as a consequence:
\begin{tcolorbox}[error, title={\small\textbf{Output with Factual Hallucination}}]
    \{\\
        \hspace*{1.5em} "Start\_Date": \{"present": \textbf{true}, "value": \textbf{"20111014"}\}\\
        \hspace*{1.5em} "End\_Date": \{"present": true, "value": "20111015"\}\\
    \}
\end{tcolorbox}
These two examples illustrate distinct modes of factual hallucination. In the \emph{deduction case}, the pipeline invents a new numerical value so that the arithmetic matches the stated total. In the \emph{date case}, the pipeline fabricates a missing field by repurposing another field. In both scenarios, the error arises not from incorrect arithmetic or logical reasoning but from introducing evidence that is absent from the source document.

\subsubsection{Numerical Infidelity Rate (NIR).}

Given a fixed rule $r$ with evaluation set 
$\mathcal{D}_r=\{d_1,\dots,d_{n_r}\}$,
let $\hat{\varphi}_{i,r}$ denote the predicted formula or reasoning process appearing in the model’s output explanation trace for $(d_i,r)$,
and $v^\star_{i,r}$ the gold numerical result.

We mark a sample $(d_i,r)$ as numerically infidel if the evaluated 
result of the predicted formula or reasoning process deviates from the gold value:
\begin{equation}
\label{eq:ni-sample}
\mathrm{NI}_{i,r} = 
\mathbf{1}\!\left[
  \mathrm{eval}(\hat{\varphi}_{i,r})
  \neq v^\star_{i,r}
\right],
\end{equation}
with deviations judged within absolute/relative tolerances.

The rule-level rate is
\begin{equation}
\label{eq:nir-rule}
\mathrm{NIR}_{r}
= \frac{1}{|\mathcal{D}_r|}
  \sum_{d_i\in\mathcal{D}_r}\mathrm{NI}_{i,r},
\end{equation}
and the group-level average for $C\subseteq\mathcal{R}$ is
\begin{equation}
\label{eq:nir-group-w}
\mathrm{NIR}^{\mathrm{w}}_{C}
= \frac{\sum_{r\in C}|\mathcal{D}_r|\cdot \mathrm{NIR}_{r}}
       {\sum_{r\in C}|\mathcal{D}_r|}.
\end{equation}

Note that hallucinations in formula generation are already captured under FHR; 
NIR isolates purely numerical inconsistencies after formula or reasoning process generation.

\paragraph{Example.}
Using the same rule for the YTD deduction value as in ~\ref{app:metrics_FHR}:
\begin{tcolorbox}[error, title={\small\textbf{Output with Numerical Infidelity}}]
    \{\\
        \hspace*{1.5em} "YTD\_Total\_Deduction": \{\\
        \hspace*{3em}"calculation": "1427.28 + 4683.88 + 344.28 + 660.50 + 617.50 = \textbf{7633.44}",\\
        \hspace*{3em}"stated": "7633.44",\\
        \hspace*{3em}"equal": true\\
        \hspace*{1.5em}\},\\
    \}
\end{tcolorbox}
Here, all evidence terms are faithfully extracted, and the pipeline correctly generates the formula. But the result is miscomputed as $7633.44$ instead of the correct $7733.44$, possibly influenced by the stated value in the document. Unlike factual hallucination, this error does not arise from fabricating or misattributing values, but from applying the correct evidence while failing to maintain consistency in quantitative reasoning and arithmetic derivations.

\subsubsection{Edge Case Handling (ECH).}
For each rule $r$ with evaluation set $\mathcal{D}_r$,
we predefine a subset $\mathcal{D}^{\mathrm{edge}}_r \subseteq \mathcal{D}_r$
containing complex or exception-driven cases
(typically 10\%--25\% of document--rule pairs).
Such cases include missing or abnormal values,
atypical field combinations, and boundary conditions
requiring adaptive reasoning beyond commom patterns.

For a document $d_i \in \mathcal{D}^{\mathrm{edge}}_r$, 
let $\hat{y}_{i,r}\in\{\texttt{Pass},\texttt{Fail}\}$ be the model's predicted label
and $y^\star_{i,r}$ the gold label. We define the indicator
\begin{equation}
\label{eq:ech-sample}
\mathrm{EC}_{i,r} = 
\begin{cases}
1, & \hat{y}_{i,r}\neq y^\star_{i,r}, \\[4pt]
0, & \hat{y}_{i,r}= y^\star_{i,r}.
\end{cases}
\end{equation}

The rule-level error rate is
\begin{equation}
\label{eq:ech-rule}
\mathrm{ECH}_{r}
= \frac{1}{|\mathcal{D}^{\mathrm{edge}}_r|}
  \sum_{d_i\in \mathcal{D}^{\mathrm{edge}}_r}\mathrm{EC}_{i,r},
\end{equation}
and the group-level average is
\begin{equation}
\label{eq:ech-group}
\mathrm{ECH}^{\mathrm{w}}_{C}
= \frac{\sum_{r\in C}|\mathcal{D}^{\mathrm{edge}}_r|\cdot \mathrm{ECH}_{r}}
       {\sum_{r\in C}|\mathcal{D}^{\mathrm{edge}}_r|}.
\end{equation}

Higher values of $\mathrm{ECH}$ indicate weaker generalization
and lower robustness on edge conditions.

\paragraph{Example.}
Consider the rule: “Is the current gross pay calculated correctly by summing up the current amount of all earning items?”
In the sample (Figure~\ref{fig:paystub_sample}), the earnings table contains a subtotal of $1663.44$ followed by a negative adjustment of $38.44$. Correct handling requires the pipeline to exclude the subtotal from the formula and include the adjustment as a subtraction. The ground truth is:
\begin{tcolorbox}[ground_truth]
    \{\\
        \hspace*{1.5em} "Current\_Gross\_Pay": \{\\
        \hspace*{3em}"calculation": "1625.00 + 32.50 + 5.94 \textbf{- 38.44} = 1625.00",\\
        \hspace*{3em}"stated": "1625.00",\\
        \hspace*{3em}"equal": true\\
        \hspace*{1.5em}\}\\
    \}
\end{tcolorbox}
However, the pipeline misinterprets the table:
\begin{tcolorbox}[error, title={\small\textbf{Output with Edge Case Error}}]
    \{\\
        \hspace*{1.5em} "Current\_Gross\_Pay": \{\\
        \hspace*{3em}"calculation": "1625.00 + 32.50 + 5.94 \textbf{+ 1663.44 + 38.44} = 3365.32",\\
        \hspace*{3em}"stated": "1625.00",\\
        \hspace*{3em}"equal": false\\
        \hspace*{1.5em}\},\\
    \}
\end{tcolorbox}
In this output, the pipeline incorrectly reuses the subtotal $1663.44$ as if it were another earning item, and also flips the negative adjustment $38.44$ into a positive contribution. The error is not due to arithmetic mistakes but to misinterpreting atypical structures in the table. Unlike factual hallucination (fabricating values) or numerical infidelity (miscomputing a correct formula), this case reflects weaker robustness to edge conditions. Instead of applying rules mechanically, a reliable industrial pipeline should capture the underlying business logic and correctly adapt it to the diverse formats and exception cases present in financial documents.

\clearpage
\twocolumn[
\refstepcounter{subsection}
\label{sec:dataset}
\subsectionmark{Evaluation Dataset}
{\bfseries \large \thesubsection\quad Evaluation Dataset}\par
\addcontentsline{toc}{subsection}{\protect\numberline{\thesubsection}Evaluation Dataset}

\noindent
Our evaluation uses a proprietary collection of production-level mortgage application documents from an active industrial workflow. The corpus spans a broad range of core and supporting document categories---covering proof of income, property appraisal, account statements, tax forms, and legal agreements---with file lengths varying from single-page forms to multi-dozen-page reports. Unlike public datasets that are often synthetic or visually uniform, our corpus retains the full heterogeneity, layout irregularities, OCR noise, and compliance constraints encountered in real underwriting. While individual files and categories vary across evaluations, the overall data composition and associated challenges are fully described, enabling reproducibility on comparable financial document collections.

\vspace{0.8em}

\begin{center}
\footnotesize
\captionof{table}{Representative mortgage document types and key characteristics.}
\label{table:dataset}

\rowcolors{2}{white}{rowgray}
\begin{tabularx}{\textwidth}{p{3.3cm} Y Y p{2.1cm}}
\toprule
\textbf{Document Type} & \textbf{Key Structural Features} & \textbf{Key Semantic Features} & \textbf{Valid Rule Categories} \\
\midrule
Appraisal &
Multi-page PDF; photos + valuation tables; firm templates; checkboxes &
Market value; comparable properties; adjustments; effective date &
$C_1$, $C_2$ \\

\seqsplit{Bank\ Account\ Statement} &
Multi-column ledgers; footnotes inside tables; embedded logos and stamps &
Balance progression; transaction category; cross-month reconciliation &
$C_1$ – $C_5$ \\

Employment\ Letter &
Letterhead; signatures; uneven paragraphs; low-contrast scans &
Title; start date; compensation; pay frequency vs. paystubs &
$C_1$, $C_2$, $C_3$ \\


\seqsplit{Investment\ Account\ Statement} &
Dense holdings tables; various formats; small-font disclosures and disclaimers &
Issue/maturity date; interest rate; holding list; balance &
$C_1$, $C_3$, $C_4$, $C_5$ \\

\seqsplit{Mortgage\ Statement} &
Multi-section layout; various formats; tables with merged cells &
Outstanding balance; interest rate and type; payment due date; amount due &
$C_1$, $C_3$, $C_4$, $C_5$ \\

\seqsplit{Notice\ of\ Assessment}&
CRA form; dense numeric blocks; tiny labels; presence of official headers &
Total/taxable income; refund amount; social insurance number; match T4 &
$C_1$, $C_2$, $C_3$, $C_4$ \\

Paystub &
Multiple tables and forms; various layouts; faded scans &
Gross; net; deductions; start/end/pay date &
$C_1$ - $C_5$ \\

\seqsplit{Personal\ Income\ Tax\ Return}&
Multipage; mixed sections; checkboxes; pre-filled and handwritten; cross-page linkage &
Address; citizenship; total/employment/net income &
$C_1$, $C_2$, $C_4$ \\



\seqsplit{Property\ Tax\ Bill} &
Multiple tables; various layouts; inconsistent labelling for tax components; watermarks and logos &
Annual tax; installments/penalties; roll/assessment IDs; schedule math &
$C_1$ - $C_5$ \\

Purchase\ Sales\ Agreement &
Long contract; initials/signatures; clause order varies; appendices with separate numbering &
Property address; price; closing date; buyer/seller name &
$C_1$, $C_2$, $C_3$ \\

Realtor\ Listing &
Embedded images; dense tables and forms; inconsistent field ordering &
List price; property address; listing date and status; square footage &
$C_1$, $C_2$ \\

\seqsplit{Realtor\ Listing\ UW} &
Listing + UW notes; handwritten overlays &
Adjusted price; remarks; reconcile with appraisal/APS &
$C_1$, $C_2$, $C_3$ \\

\seqsplit{Rental\ Lease\ Agreement} &
Multi-page legal contract; various templates; small-font clauses; handwritten &
Monthly rent; term dates; payment plan; residence overlap &
$C_1$, $C_2$, $C_3$ \\

\seqsplit{Separation\ Agreement} &
Multi-page legal document; dense narrative clauses; annex schedules &
Support obligations; asset division; enforceable clauses; child custody &
$C_1$, $C_2$, $C_3$ \\


\seqsplit{T4:\ Statement\ of\ Remuneration\ Paid} &
Fixed CRA box layout; small-font numeric fields; bilingual field labels; scan noise &
Employment income; Province; EI/CPP exemptions; align paystub/NOA &
$C_1$, $C_2$, $C_4$ \\

\seqsplit{T4A:\ Statement\ of\ Pension,\ Retirement,\ Annuity,\ and\ Other\ Income} &
Fixed CRA box layout; small-font numeric fields; bilingual field labels; scan noise &
Pension/annuity/self-employment income; income tax deducted; social insurance number  &
$C_1$, $C_2$, $C_4$ \\


\seqsplit{T4A(P):\ Statement\ of\ Canada\ Pension\ Plan\ Benefits} &
Fixed CRA box layout; small-font numeric fields; bilingual field labels; scan noise &
CPP benefits amount; Income tax deducted;  &
$C_1$, $C_2$, $C_4$ \\

\seqsplit{T4RIF:\ Statement\ of\ Income\ From\ a\ Registered\ Retirement\ Income\ Fund} &
Fixed CRA box layout; small-font numeric fields; bilingual field labels; scan noise &
Taxable amounts; Payer/issuer name; taxable amount; social insurance number &
$C_1$, $C_2$, $C_4$ \\

\seqsplit{T5:\ Statement\ of\ Investment\ Income} &
Fixed CRA box layout; small-font numeric fields; bilingual field labels; scan noise &
Actual amount of dividends; Interest from Canadian sources; Reported income vs. bank statements &
$C_1$, $C_2$, $C_5$ \\

\seqsplit{T776:\ Statement\ of\ Real\ Estate\ Rentals} &
Fixed CRA box layout; small-font numeric fields; bilingual field labels; scan noise &
Gross rents; Total expenses; Net income (loss) before adjustments; Capital cost allowance (CCA) claim &
$C_1$ - $C_5$ \\
\bottomrule
\end{tabularx}

\rowcolors{0}{}{} 
\end{center}
\vspace{-0.5em}
]



\end{document}